\documentclass{article}

\PassOptionsToPackage{numbers}{natbib}

     \usepackage[final]{neurips_2019}

\usepackage[utf8]{inputenc} 
\usepackage[T1]{fontenc}    
\usepackage{hyperref}       
\usepackage{url}            
\usepackage{booktabs}       
\usepackage{amsfonts}       
\usepackage{nicefrac}       
\usepackage{microtype}      

\usepackage{enumitem}

\usepackage{graphicx}
\usepackage{epstopdf}
\usepackage{subfig}

\usepackage{todonotes}

\usepackage{amsmath}
\usepackage{tikz}
\usepackage{mathdots}
\usepackage{yhmath}
\usepackage{cancel}
\usepackage{color}
\usepackage{siunitx}
\usepackage{array}
\usepackage{multirow}
\usepackage{amssymb}
\usepackage{gensymb}
\usepackage{tabularx}
\usepackage{booktabs}
\usetikzlibrary{fadings}

\title{Avoiding hashing and encouraging visual semantics in referential emergent language games}

\author{%
  Daniela Mihai\thanks{Vision, Learning and Control Group, Electronics and Computer Science, University of Southampton, \{adm1g15, jsh2\}@ecs.soton.ac.uk}\\
  \And
  Jonathon Hare\footnotemark[1]\\
}

\begin{document}

\maketitle

\begin{abstract}

There has been an increasing interest in the area of emergent communication between agents which learn to play referential signalling games with realistic images. In this work we consider the signalling game setting of~\citeauthor{Havrylov2017} and investigate the effect of the feature extractor's weights and of the task being solved on the visual semantics learned or captured by the models. We impose various augmentation to the input images and additional tasks in the game with the aim to induce visual representations which capture conceptual properties of images. Through our set of experiments, we demonstrate that communication systems which capture visual semantics can be learned in a completely self-supervised manner by playing the right types of game.
\end{abstract}

\section{Introduction}
The idea that agents might learn language by playing visually grounded games has a long history~\cite{cangelosi2002simulating, steels2012experiments}.  Research in this space has recently had something of a resurgence with the introduction of a number of models that simulate the play of \textit{referential} games~\cite{Lewis1969-LEWCAP-4} using realistic visual inputs~\cite{lazaridou2017multi,Havrylov2017,lee2017emergent}. On one hand, these works have shown that the agents can learn to successfully communicate to play these games; however, on the other hand, there has been much discussion as to whether the agents are really learning a semantically visually grounded communication system. The recent paper by \citet{BouchacourtB18} highlights this issue in the context of a pair of games designed by~\citet{lazaridou2017multi} which involved the sender and receiver agents being presented with pairs of images. 

The work of \citet{Havrylov2017} explored a different, and potentially harder, game than those proposed by~\citet{lazaridou2017multi}. In their work, they show some qualitative examples in which it does appear that the generated language does in some way convey the visual semantics of the scene. There are however many open questions from this analysis. One of the key questions is to what extent the ImageNet-pretrained VGG-16 CNN \cite{simonyan2014very} used in the model is affecting the language that emerges.

In this work, we explore visual semantics in the context of \citet{Havrylov2017}'s game by carefully controlling the visual feature extractor that is used and augmenting the game play in different ways. We seek to explore what factors encourage the emergent language to convey visual semantics rather than falling back to a communication system that just learns hashes of the input images. More concretely, we:
\begin{itemize}
    \item explore the effect of different weights in the CNN used to generate the features (fixed, pretrained on ImageNet and frozen (as in the original work), random and frozen, and, learned end-to-end in the model;
    \item explore the effect of augmentations that make the game harder by changing the image given to the sender (adding noise and/or randomly rotating), but not the receiver;
    \item extend the game to include a secondary task (guessing the rotation of the sender's input) in order to assess whether having agents perform more diverse tasks might lead to stronger visual semantics emerging.
\end{itemize}
We draw attention to the fact that other than in the cases where we use pretrained feature extractors, our simulations are completely self-supervised, and there is no explicit signal of what a human would understand as the `visual semantics' given to the models at any point.

\section{Model and Experiments}

\begin{figure}
    \centering
    \resizebox{0.6\textwidth}{!}{\input{model-orig.tikz}\unskip}
    \caption{\citet{Havrylov2017}'s game setup and model architecture.}
    \label{fig:emergentlanggame}
\end{figure}    

\citet{Havrylov2017}'s model and game are illustrated in Figure~\ref{fig:emergentlanggame}. The objective of the game is for the Sender agent to communicate information about an image it has been given to allow the Receiver agent to correctly pick the image from a set containing many (127 in all experiments) distractor images. The Sender agent utilises an LSTM to generate a sequence of tokens given a hidden state initialised with visual information and a Start of Sequence (SoS) token. To ensure that a sequence of only discrete tokens is transmitted, the output token logits produced by the LSTM cell at each timestep are sampled with the Straight-Through Gumbel Softmax operator (GS-ST).\footnote{\citet{Havrylov2017} experimented with ST-GS, the relaxed Gumbel Softmax and REINFORCE in their work, however, we focus our attention on ST-GS here.} The ST-GS provides a one-hot vector at each time step but uses the gradients of the relaxed Gumbel Softmax during the backward pass to circumvent the problem that the sampling is non-differentiable~\cite{DBLP:conf/iclr/MaddisonMT17,DBLP:conf/iclr/JangGP17}. The Receiver agent uses an LSTM to decode the sequence of tokens produced by the Sender, from which the output is projected into a space that allows the Receiver's image vectors to be compared using a dot product. \citet{Havrylov2017} use a fixed VGG16 CNN pretrained on ImageNet to extract image features in both agents. The model is trained using a hinge-loss objective to maximise the probability of the correct image being chosen. The Sender can generate messages up to a given maximum length; shorter codes are generated by the use of an end of sequence token. Although not mentioned in the original paper, we found that the insertion of a BatchNorm layer in the Sender between the CNN and LSTM, and after the LSTM in the Reciever, was critical for learnability and reproduction of the original experimental results.   

\paragraph{Experimental setup}
Our experiments use the model described above with some modifications under different experimental settings. In all cases, we perform experiments using the CIFAR-10 dataset rather than the COCO dataset used in the original work (to replicate the original results requires multiple GPUs due to the memory needed, as well as considerable training time). In light of the smaller resolution images and lower diversity of class information, we choose a word embedding dimension of 64, hidden state dimension of 128, and total vocabulary size of 100 (including the EoS token). We also limit the maximum message length to 5 tokens. The training data is augmented using color jitter ($p_{bri}=0.1, p_{con}=0.1, p_{sat}=0.1, p_{hue}=0.1$), random grayscale transformation ($p=0.1$), and random horizontal flipping ($p=0.5$), so there is very low probability of the model seeing the same image more than once. The batch size is set to 128, allowing for the Receiver to see features from the target image plus 127 distractors. Most simulations converge or only slowly improve after about 60 epochs, however for consistency, all results are reported on models trained to 200 epochs where convergence was observed to be guaranteed. 

\paragraph{Metrics}
Our key objective is to measure how much visual semantic information is being captured by the emergent language. If humans were to play this game, it is clear that a sensible strategy would be to describe the target image by its semantic content (e.g. "a yellow car front-on" in the case of the example in Figure~\ref{fig:emergentlanggame}). It is also reasonable to assume in the absence of strong knowledge about the make-up of the dataset (for example, that the colour yellow is relatively rare) that a semantic description of the object in the image (a ``car'') should have a strong part to play in the communicated message if visual semantics are captured. In the case of CIFAR-10 dataset in which most images have a single subject, ``objectness'' can be considered a measure of semantics. With this in mind, we can measure to what extent the communicated messages capture the object by looking at how the target class places in the ranked list of images produced by the Receiver. More specifically, in the top-5 ranked images guessed by the Receiver, we can calculate the number of times the target object class appears, and across all the images we can compute the average of the ranks of the images with the matching class. In the former case, if the model captures more semantic information, the number will increase; in the latter, the mean-rank decreases if the model captures more semantic information. A model (which is successful at communicating) which performs almost ideal hashing would have an expected top-5 number of the target class approaching 1.0 and expected average rank of 59.5, whilst a model that completely captures the ``objectness'' (and still guesses the correct image) would have an expected top-5 target class count of 5, and mean rank of between 6.5 and 7. In addition to these metrics for measuring visual semantics, we also measure top-1 and top-5 communication success rate (Receiver guesses correctly in the top-1 and top-5 positions) and the message length for each trial.

\paragraph{The effect of different weights in the visual feature extractor.}
Generating and communicating hash codes is very clearly an optimal (if very unhuman) way to play the image guessing game successfully, however in ~\citet{Havrylov2017}'s original work there was qualitative evidence that did not happen when the model was trained, and that visual semantics were captured. To what extent is this caused by the pretrained feature extractor? We attempt to answer this question by exploring three different model variants: the original model with the CNN fixed and initialised with ImageNet weights; the CNN fixed, but initialised randomly; and, the CNN initialised randomly, but allowed to update its weights during training. 
Results from these experiments are summarised in Table~\ref{table:weights}. The first observation relates to the visual-semantics measures; it is clear (and unsurprising) that the pretrained model captures the most semantics of all the models. It is also reasonable that we observe less semantic alignment with the end-to-end model; without external biases, this model should be expected to move towards a hashing solution. It is somewhat surprising however that the end-to-end model doesn't have the best communication success rate, and is outperformed by the random model. It is already known that a randomly initialised CNN can provide reasonable features~\cite{Saxe:2011:RWU:3104482.3104619}; during training, the Sender and Receiver convergence had particularly low variance with this model, allowing the agents to quickly evolve a successful strategy. One might question if the end-to-end model was handicapped because it had more weights to learn in the same number of epochs, however training all the models for 1000 epochs yielded only a $2\%$ improvement in communication rate across the board.


\begin{table}
  \caption{The effect of different weights in the feature extractor CNN. Measures are averaged across 10000 games on the CIFAR-10 validation set using a single trained model. Values in brackets are standard deviations across games.}
  \label{table:weights}
  \centering
   \begin{tabular}{llllll}
    \toprule
     Feature extractor & Comm. & Message & Top-5 & \#target-class  & Target-class \\
       & rate & length & comm. rate & in top-5 & avg. rank \\
    \midrule
     Pretrained \& fixed &0.88 ($\pm$0.31) &4.91 ($\pm$0.38) &0.99 &1.84 &46.58\\
 
     Random \& frozen &0.95 ($\pm$0.19) &4.96 ($\pm$0.23) &1 &1.66 &52.13\\
 
    Learned end-end &0.89 ($\pm$0.3) &4.90 ($\pm$0.35) &1 &1.53 &54.01\\
    \bottomrule
  \end{tabular}
\end{table}

\paragraph{Making the game harder with augmentation.}
We next investigate the behaviour of the same three model variants while playing a slightly more difficult game. The input images to the Sender are augmented with additional noise and random rotations, and thus will not be pixel-identical with those seen by the Receiver. For the model to communicate well it must either capture the semantics or learn to generate highly-robust hash codes. The added noise is generated from a normal distribution with mean 0 and variance 0.1, and the rotations applied to the input images are randomly chosen from \{\ang{0}, \ang{90}, \ang{180}, \ang{270}\}. 

Table~\ref{table:noiserot} shows the effect of adding noise and rotations. In general noise results in a slight increase in the communication success rate. More interestingly, for randomly rotated Sender images the augmentation tends to increase the visual semantics captured by all the models, although this is most noticeable in the pretrained variant. At the same time, the communication success rate of the pretrained model drops; future experiments will quantify this effect for longer message lengths.

\begin{table}
  \caption{The effect of different weights in the feature extractor CNN when the model is augmented by adding noise or random rotations to the Sender agent's input images. Measures as per Table~\ref{table:weights}.}
  \label{table:noiserot}
  \centering
   \begin{tabular}{llllll}
    \toprule
     Feature extractor & Comm. & Message & Top-5 & \#target-class  & Target-class \\
       & rate & length & comm. rate & in top-5 & avg. rank \\
    \midrule
    \multicolumn{6}{l}{\textbf{Sender images augmented with Gaussian noise:}}\\
      Pretrained \& fixed & 0.92 ($\pm$0.26) & 4.97 ($\pm$0.21) & 0.99 & 1.85 & 46.12\\

      Random \& frozen & 0.96 ($\pm$0.19) & 4.92 ($\pm$0.34) & 1 & 1.6 & 54.01\\

      Learned end-end & 0.94 ($\pm$0.23) & 4.92 ($\pm$0.33) & 1 & 1.5 & 57.11\\
    \midrule
    \multicolumn{6}{l}{\textbf{Sender images augmented with random rotations:}}\\
      Pretrained \& fixed & 0.83 ($\pm$0.37) & 4.99 ($\pm$0.14) & 0.99 &2.07 &41.89\\

      Random \& frozen & 0.87 ($\pm$0.33) & 4.91 ($\pm$0.38) & 0.99 &1.7 &51.89\\

      Learned end-end & 0.92 ($\pm$0.25) & 4.93 ($\pm$0.33) & 1 & 1.6 &55.96\\
    \bottomrule
  \end{tabular}
\end{table}





\paragraph{Can a pair of agents learn semantic visual representations without supervision?} The experimental results above clearly show that the fully-learned models always collapse towards solutions not aligned with human notations of visual semantics. Conversely, the use of a network that was pretrained in a supervised fashion to classify real-world images has a positive effect on the ability of the communication system to capture visual semantics. We end by exploring if it might be possible for a communications protocol with notions of visual semantics to emerge directly from pure self-supervised game-play. In order to achieve this, we propose that the agents should not only learn to play the referential game, but they should also be able to play other games (or solve other tasks). In our initial experiments we formulate a setup where the agents not only have to play the augmented version of the game described above (with both noise and rotations randomly applied to the image given to the Sender, but not the Receiver), but also one of the agents has to guess the rotation of the image given to the Sender (see Appendix~\ref{app:rotmodels}). This choice of the additional task is motivated by \citet{gidaris2018unsupervised} who showed that a self-supervised rotation prediction task could lead to good features for transfer learning, on the premise that in order to predict rotation the model needed to recognise the object. Results of these experiments are shown in Table~\ref{table:extmodels}. Whilst there is still a way to go to achieve the same levels of game-play performance shown in the Tables~\ref{table:weights} and~\ref{table:noiserot}, it is clear that these fully self-supervised end-to-end trained models can both learn a communication system to play the game(s) that diverges from a hashing solution towards something that better captures semantics. 

\begin{table}
  \caption{End-to-End learned models with an additional rotation prediction task. Measures as per Table~\ref{table:weights}, except for the inclusion of the accuracy of rotation prediction.}
  \label{table:extmodels}
  \centering
   \begin{tabular}{llllll}
    \toprule
     Model & Comm. & Top-5 & \#target-class  & Target-class & Rot.\\
       & rate & comm. rate & in top-5 & avg. rank & acc.\\
    \midrule
      Receiver-Predicts (Fig.~\ref{fig:emergentlanggame-rot}) & 0.64 ($\pm$0.48) & 0.95 & 1.84 & 45.6 & 0.82\\
      Sender-Predicts (Fig.~\ref{fig:emergentlanggame-srot}) & 0.69 ($\pm$0.46) & 0.98 & 2.05 & 43.41 & 0.84\\
    \bottomrule
  \end{tabular}
\end{table}




\section{Conclusion}
In this work, we first quantify the effect that using a pretrained visual feature extractor has on the ability of the language that emerges from a pair of agents learning to play a referential game to capture visual semantics. We demonstrate that it is possible to formulate a multiple-game setting in which the emergent language is \textit{more} semantically grounded without the need for any outside supervision. We note these models represent difficult multi-task learning problems, and that the next steps in this direction would benefit from full consideration of recent approaches to multi-task learning which deal with multiple objectives that conflict~\cite{SenerNips2018}.

\bibliographystyle{plainnat}
\bibliography{refs}

\appendix
\renewcommand\thefigure{\thesection.\arabic{figure}}
\setcounter{figure}{0}

\section{Extended Model Configuration}
\label{app:rotmodels}

\begin{figure}[h]
    \centering
    \resizebox{0.7\textwidth}{!}{\input{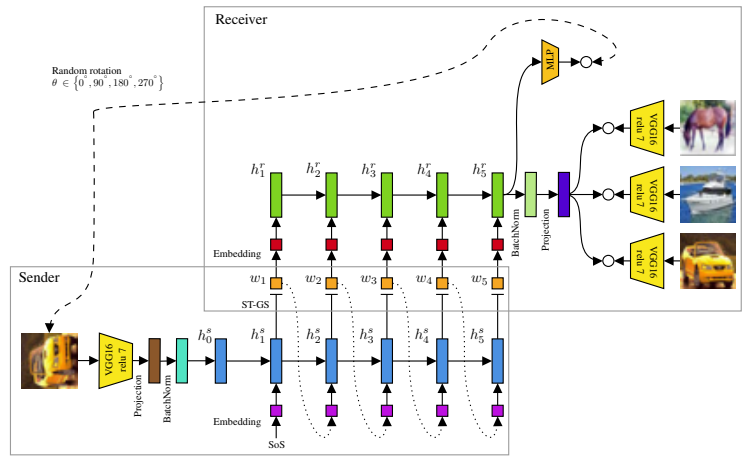}\unskip}
    \caption{Extended game with the Receiver also required to guess the orientation of the Sender's image.}
    \label{fig:emergentlanggame-rot}
\end{figure}

\begin{figure}[h]
    \centering
    \resizebox{0.7\textwidth}{!}{\input{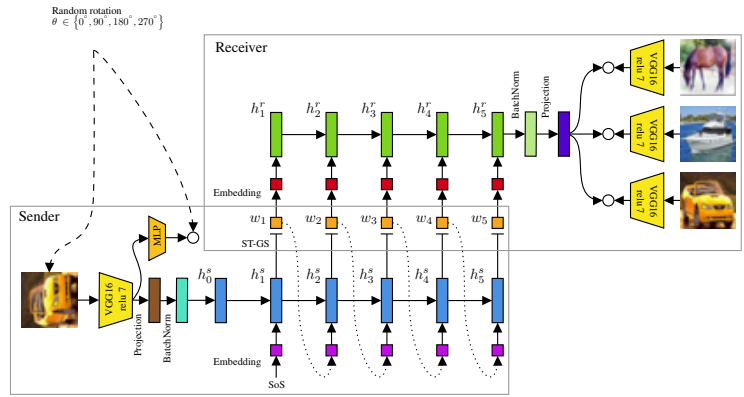}\unskip}
    \caption{Extended game with the Sender augmented with an additional loss based on predicting the orientation of the input image.}
    \label{fig:emergentlanggame-srot}
\end{figure}

Our extended dual task models are shown in Figures~\ref{fig:emergentlanggame-rot} and~\ref{fig:emergentlanggame-srot}. The only difference from the model described in the main body is the addition of the rotation prediction network. The rotation prediction network is inspired by~\citet{gidaris2018unsupervised}, and consists of three linear layers with Batch Normalisation before the activation functions. The first two layers use ReLU activations, and the final layer uses a Softmax to predict the probability of the four possible rotation classes. With the exception of the final layer, each layer outputs 200-dimensional vectors. Cross Entropy is used as the loss function for the rotation prediction task ($\mathcal{L}_{rotation}$). All other model parameters and the game-loss definition match those described in the main body of the paper.

Training these models is harder than the original Sender-Receiver model because the gradients pull the visual feature extractor in different directions --- the game achieves good performance when the features behave like hash codes, whereas the rotation prediction task requires much more structured features. This conflict means that it is difficult to train the models such that they have the ability to solve both tasks. For the experiments in the paper, the results for the Sender-predicts model (Figure~\ref{fig:emergentlanggame-srot}) used a weighted addition $0.5\cdot\mathcal{L}_{rotation} + \mathcal{L}_{game}$, where $\mathcal{L}_{game}$ refers to the original hinge-loss objective for the game proposed by \citet{Havrylov2017}. For the Receiver-predicts model (Figure~\ref{fig:emergentlanggame-rot}) the results were obtained by switching between optimising $5.0\cdot\mathcal{L}_{rotation}$ and $5.0\cdot\mathcal{L}_{rotation} + \mathcal{L}_{game}$ on alternate batch iterations. For the latter model we also tried using additive loss with learned weights (following~\citet{kendall2017multi}) however this created a model with good game-play performance, but an in-ability to predict rotation (and poor semantic representation ability). Clearly further work in developing optimisation strategies for these multi-game models is of critical importance in future work.

\end{document}